\pdfoutput=1
\documentclass[11pt]{article}

\usepackage{EACL2023}

\usepackage{times}
\usepackage{latexsym}
\usepackage{hyperref}
\usepackage{graphicx}
\usepackage{makecell}
\usepackage{tabularx}
\usepackage{booktabs}
\usepackage[subrefformat=parens]{subcaption}

\usepackage[T1]{fontenc}
\usepackage[utf8]{inputenc}

\usepackage{microtype}

\usepackage{inconsolata}
\usepackage{amssymb}
\usepackage{amsmath}
\usepackage{multirow}

\title{Entity-Aware Dual Co-Attention Network for Fake News Detection}

\author{Sin-Han Yang $^1$, Chung-Chi Chen $^2$, Hen-Hsen Huang $^3$, Hsin-Hsi Chen $^{1}$\\
  $^1$ Department of Computer Science and Information Engineering, \\ National Taiwan University, Taiwan\\
  $^2$ AIST, Japan\\
  $^3$ Institute of Information Science, Academia Sinica, Taiwan\\
\texttt{b08202029@ntu.edu.tw, c.c.chen@acm.org} \\
\texttt{hhhuang@iis.sinica.edu.tw, hhchen@ntu.edu.tw}\\}

\begin{document}
\maketitle

\begin{abstract}
Fake news and misinformation spread rapidly on the Internet. 
How to identify it and how to interpret the identification results have become important issues. 
In this paper, we propose a Dual Co-Attention Network (Dual-CAN) for fake news detection, which takes news content, social media replies, and external knowledge into consideration. 
Our experimental results support that the proposed Dual-CAN outperforms current representative models in two benchmark datasets. 
We further make in-depth discussions by comparing how models work in both datasets with empirical analysis of attention weights.\footnote{Code repository: \url{https://github.com/SinHanYang/Dual-CAN}}
\end{abstract}

\section{Introduction}
The development of the Web and social media platforms helps us obtain news quickly, but also provides a gateway for spreading false information. 
The impact of false information is wide, and the spread speed might be even faster than the actual one~\cite{vosoughi2018spread}. 
For example, fake news is proven empirically to influence the 2016 U.S. presidential election~\cite{bovet2019influence,grinberg2019fake,budak2019happened}.
Given the impact of false information, previous studies paid a lot of effort to detect it from different aspects, including (1) news content only~\cite{santos-etal-2020-measuring,kim-ko-2021-graph}, (2) the combination of news articles and social media replies~\cite{li-etal-2020-exploiting,lu-li-2020-gcan}, and (3) additional publisher/user information~\cite{long-etal-2017-fake,yuan-etal-2020-early,del-tredici-fernandez-2020-words}.
In this work, we focus on using both news contents and social media replies, and further add external knowledge to enhance the model's ability to capture critical entities. 

% \begin{figure}
%     \centering
%     \includegraphics[width=0.5\textwidth]{vsprevious_v3.png}
%     \caption{Comparison between the proposed Dual-CAN and previous methods.}
%     \label{fig:vsprevious}
% \end{figure}

Named entities play an important role in document understanding and influence text generation performances~\cite{narayan2021planning,narayan-etal-2022-well}.
Inspired by this notion, we design a novel model, named Dual Co-Attention Network (Dual-CAN), which takes entities' descriptions into consideration to enhance the background knowledge of the model. 
%Figure~\ref{fig:vsprevious} provides a comparison between the proposed Dual-CAN and the representative model dEFEND~\cite{10.1145/3357384.3357862} in fake news detection. 
The proposed Dual-CAN is modified based on one of the representative fake news detection models, dEFEND~\cite{shu2019defend}.
There are three major improvements in the proposed Dual-CAN: 
(1) Inspired by \citet{hu2021compare}, we add entities' descriptions for enhancing the performance.
(2)  Instead of using LSTM-based architectures \cite{shu2019defend,lu-li-2020-gcan}, we adopt attention architecture~\cite{vaswani2017attention} as the backbone. 
(3) We further tailor-made a co-attention layer for comparing the given news article with entity descriptions.
%Figure~\ref{fig:vsprevious} summarizes the three major improvements of the proposed Dual-CAN.
In sum, in addition to adopting entity descriptions from Wikipedia, we design a new architecture to fusion all information. 
Our main contribution is providing a novel model for fake news detection and pointing out a new direction for enhancing performance. 
%While improving performance, we still keep the model's explainability for illustrating the predictions. 
%We provide the details in Section~\ref{sec:Model Interpretability Evaluation}. 

\section{Related Works}
%\subsection{Fake News Detection}
%Fake news detection is an important natural language processing (NLP) application. 
Previous works in fake news detection mainly focused on two aspects: news content based and social context based. \citet{rashkin-etal-2017-truth} focus on the linguistic characteristics of the news content to detect fake news, and find that fake news often contain specific kinds of words. 
\citet{ma2016detecting} use recurrent neural networks (RNN) to learn the hidden representations from the contextual information of relevant posts over time.
\citet{monti2019fake} analyze social graph and user profile to predict fake news. 
\citet{shu2019role} find that user profile features are useful in fake news detection. \citet{shu2019defend} and \citet{lu-li-2020-gcan} use co-attention model to leverage news content and social context. Their models not only have better performance but also provide interpretability to their models.
Several works also use external knowledge to improve model's predictions. 
\citet{wang2020fake} and \citet{hu2021compare} use entity linking method to capture entity descriptions and leverage them in their models. 
Inspired by these works, we use external knowledge for entities to enhance performance, and use both news content and social media context in the proposed model.

\section{Method}
\label{sec:method}
\begin{figure}
    \centering
    \includegraphics[width=0.49\textwidth]{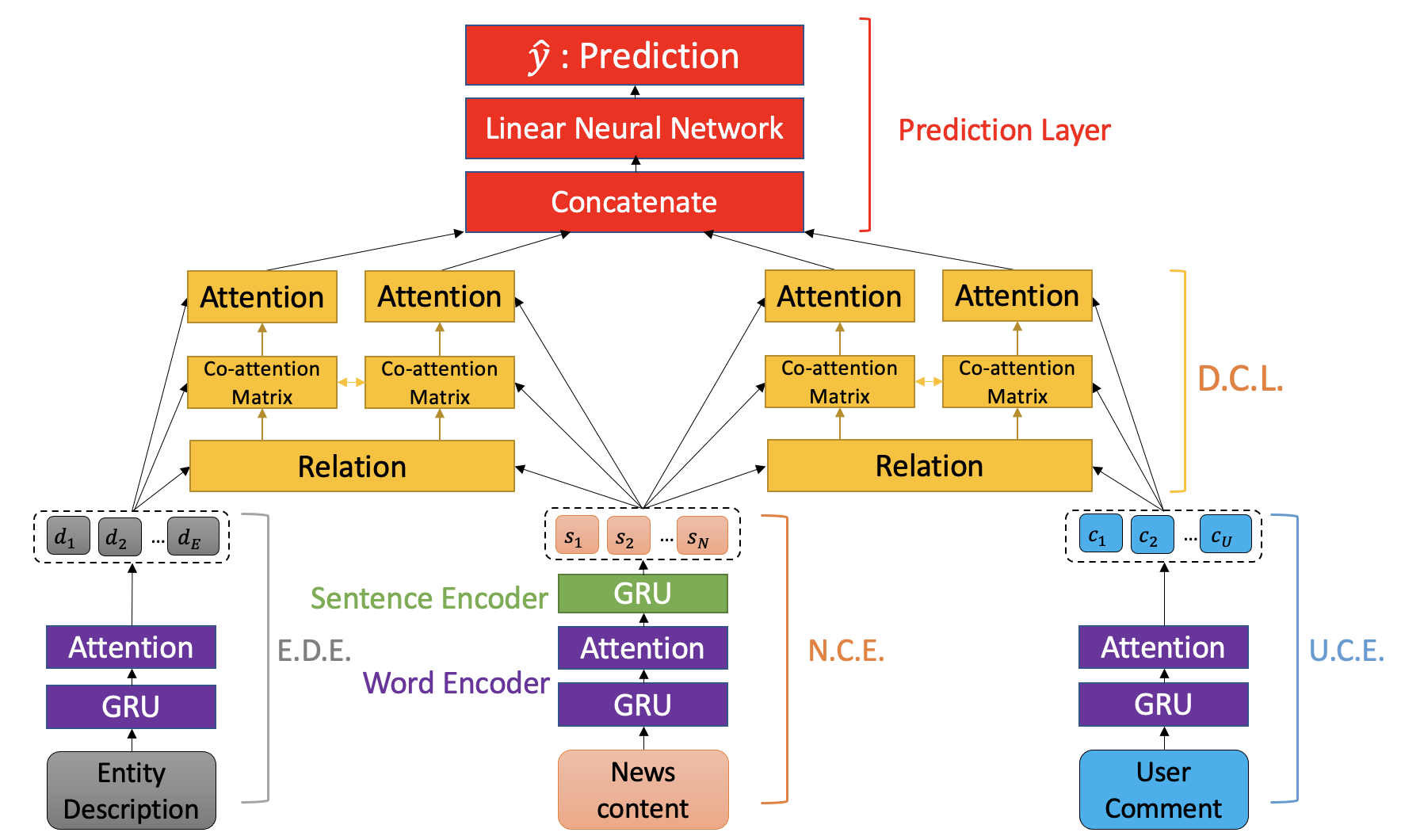}
    \caption{Architecture of Dual-CAN. D.C.L., E.D.E., N.C.E., and U.C.E. stand for dual co-attention layer, entity description encoder, news content encoder, and user comment encoder, respectively.}
    \label{fig:model_arch}
\end{figure}

Figure~\ref{fig:model_arch} shows the architecture of the proposed Dual-CAN.
This section describes the details of the proposed Dual-CAN model, which is composed of five components.\footnote{The hyperparameters are reported in Appendix \ref{sec:Implement Detail}.}
The first one is \emph{news content encoder}, which employs word-level attention network and sentence-level encoder to generate features for the corresponding news contents. 
The second is \emph{entity description encoder}. 
For each entity in news content, entity description encoder grabs its descriptions from the external knowledge base and creates features to represent them.  
The third is \emph{user engagement encoder}, which employs the same method as \emph{news content encoder} to create features to represent user comments. 
The fourth is \emph{dual co-attention component}, which  captures the relation between (news content, entity description) and (news content, user engagement) pairs. 
The last is \emph{prediction component}, which combines all information from the previous components to make the final predictions.  

\subsection{News Content Encoder}
\label{sec:4.1}
A news story is composed of a sequence of sentences $\mathbf{S}=[\mathbf{s_1},\mathbf{s_2},...,\mathbf{s_{N}}]$, and 
a sentence is composed of up to $M$ words $\mathbf{s_{i}}=[\mathbf{w_{i1}},\mathbf{w_{2i}},...,\mathbf{w_{iM}}]$. Here,
N is the maximum number of sentences in a piece of news, and M is the maximum number of words in a sentence. We perform padding to control the maximum number of sentences and words in news content. 
To create features to represent a news story, we use \emph{word-level attention network} to encode each sentence, and use \emph{sentence-level encoder} to encode all sentences in news content.

\subsubsection{Word-Level Attention Network}
\label{sec:Word-Level Attention Network}
We use Glove \cite{pennington-etal-2014-glove} to create word embedding of $d$ dimensions during the preprocessing stage for each word in sentences. 
For a sentence $\mathbf{s} \in \mathbb{R}^{d\times M}$, we utilize bi-directional Gating Recurrent Units (GRU) \cite{https://doi.org/10.48550/arxiv.1412.3555} to learn the word-level representation. 
The output of the BiGRU is $\mathbf{v_i}=\emph{BiGRU}(w_i) \in \mathbb{R}^{2h}, i\in\{1,2,...M\}$, where $h$ is the dimension of the GRU. 
Next, we perform the basic attention mechanism to increase performance and interpretability \cite{lu-li-2020-gcan} of the word encoder. Attention weight $\alpha_i$ shows the importance of the ith word. The word-level attention network generates the representation of a sentence vector $\mathbf{v'} \in R^{2h\times 1}$ calculated as follows:
\begin{equation}
    \mathbf{v'}=\sum^{M}_{i=1}\alpha_i\mathbf{v}_i
\end{equation}
where $a_i$ is:
\begin{equation}
\begin{split}
    & \mathbf{k_i}=tanh(\mathbf{P_n v_i+b_n}) \\
    & \alpha_i=\frac{exp(\mathbf{u_n k_i})}{\sum_{j=1}^{M} exp(\mathbf{u_n k_j)}}
\end{split}
\end{equation}
$\mathbf{P_n} \in \mathbb{R}^{2h\times h}$, $\mathbf{u_n} \in \mathbb{R}^{h\times 1}$ are learnable parameter. We preform a linear layer on $\mathbf{v}_i$, and use a parameter $\mathbf{k}_j$ to calculate the attention weight.

\subsubsection{Sentence-Level Encoder}
\label{sec:4.1.2}
We use BiGRU again to encode sentences in a news story. 
A sentence vector $\mathbf{s_i}\in R^{2h\times 1}$ is calculated from the output of \emph{word-level attention network}:
\begin{equation}
    \mathbf{s_i}=\emph{BiGRU}(\mathbf{v_i'}), i\in\{1,2,...,N\}
\end{equation}
Finally, single news content is represented by a list of sentence vectors $\mathbf{S}=[\mathbf{s_1},\mathbf{s_2},...,\mathbf{s_{N}}] \in \mathbb{R}^{2h\times N}$.

\subsection{Entity Description Encoder}
\label{sec:4.2}
For each news content, we identify entities in it and grab their descriptions from Wikipedia using tools TAGME \cite{10.1145/1871437.1871689}. 
For each entity description, we only use the first $E$ sentences. 
With the word-level attention network in Section \ref{sec:Word-Level Attention Network}, we create features that describe entity descriptions $\mathbf{D}=[\mathbf{d_1},\mathbf{d_2},...,\mathbf{d_{E}}]$. 
Finally, entity descriptions for a piece of news is represented by a list of sentence vectors $\mathbf{D}=[\mathbf{d_1},\mathbf{d_2},...,\mathbf{d_{E}}] \in \mathbb{R}^{2h\times E}$.

\subsection{User Comment Encoder}
For all user comments related to a news story, we only use the first $U$ sentences. 
We extract features to describe user comments  $\mathbf{C}=[\mathbf{c_1},\mathbf{c_2},...,\mathbf{c_{U}}]$ with the word-level attention network in Section~\ref{sec:Word-Level Attention Network}. 
Finally, user comments for a news story are represented by a list of sentence vectors $\mathbf{C}=[\mathbf{c_1},\mathbf{c_2},...,\mathbf{c_{U}}] \in \mathbb{R}^{2h\times U}$.

\subsection{Dual Co-Attention Component}
Because we want to know whether the entity description confirms/refutes the news content and whether user comments reflect the character of the news content, we adopt co-attention network for capturing the relationship between news content and entity descriptions, and another co-attention network for linking the relationship between news content and user comments. Given news content feature vectors $\mathbf{S}=[\mathbf{s_1},\mathbf{s_2},...,\mathbf{s_{N}}] \in \mathbb{R}^{2h\times N}$, entity description feature vectors $\mathbf{D}=[\mathbf{d_1},\mathbf{d_2},...,\mathbf{d_{E}}] \in \mathbb{R}^{2h\times E}$, and user comments feature vectors $\mathbf{C}=[\mathbf{c_1},\mathbf{c_2},...,\mathbf{c_{U}}] \in \mathbb{R}^{2h\times U}$, we use \emph{dual co-attention mechanism} for interpreting model predictions.

\subsubsection{Entity Description Co-attention}
\label{sec:4.4.1}
First, we compute a relation matrix $\mathbf{F}$
\begin{equation}
    \mathbf{F}=tanh(\mathbf{DW_rS}) \in \mathbb{R}^{E\times N}
\end{equation}
to capture the relationship between news content and entity descriptions, where $\mathbf{W_r} \in \mathbb{R}^{2h\times 2h}$ is a learnable parameter. Second, we calculate interaction maps for news content $H_s$ and entity description $H_c$,
\begin{equation}
\begin{split}
    & \mathbf{H_s}=tanh(\mathbf{W_sS}+\mathbf{W_dDF}^T) \\
    & \mathbf{H_d}=tanh(\mathbf{W_dD}+\mathbf{W_sSF})
\end{split}
\end{equation}
where $\mathbf{W_s}, \mathbf{W_d} \in \mathbb{R}^{2h\times 2h}$ are learnable parameters. Third, we calculate attention weights on each sentence in news content and entity descriptions.
\begin{equation}
\begin{split}
    & \mathbf{a_{s_1}}=softmax(\mathbf{w_{hs}H_s}) \\
    & \mathbf{a_d}=softmax(\mathbf{w_{hd}H_d})
\end{split}
\end{equation}
where $\mathbf{w_{hs}}$ and $\mathbf{w_{hd}} \in \mathbb{R}^{1\times 2h}$ are learnable parameters. After we get attention weights $\mathbf{a_{s_1}}\in \mathbb{R}^{1 \times N}, \mathbf{a_d}\in \mathbb{R}^{1 \times E}$, we generate new feature vectors for news contents and entity descriptions:
\begin{equation}
\begin{split}
    & \hat{\mathbf{s_1}}=\mathbf{a_{s_1}S}^T \\
    & \hat{\mathbf{d}}=\mathbf{a_dD}^T
\end{split}
\end{equation}
Finally, we represent news content in a feature vector $\hat{\mathbf{s_1}} \in \mathbb{R}^{1\times 2h}$, and entity descriptions in a feature vector $\hat{\mathbf{d}} \in \mathbb{R}^{1\times 2h}$.

\subsubsection{User Comment Co-attention}
We apply co-attention model as shown in Section~\ref{sec:4.4.1} to news content and user comments. 
We represent news content in a feature vector $\hat{\mathbf{s_2}} \in \mathbb{R}^{1\times 2h}$, and user comments in a feature vector $\hat{\mathbf{c}} \in \mathbb{R}^{1\times 2h}$.
The attention weights vector for news content and user comments are $\mathbf{a_{s_2}}\in \mathbb{R}^{1\times N}$ and $\mathbf{a_{c}}\in \mathbb{R}^{1\times U}$.

\begin{table*}[t]
    \centering
    \small
    \begin{tabular}{lcccccccc}
    \toprule
    \multirow{2}[2]{*}{Model (Input) (\# of Parameters)} & \multicolumn{4}{c}{GossipCop} & \multicolumn{4}{c}{CoAID}\\
     & Accuracy & F1 & Precision & Recall & PR-AUC & F1 & Precision & Recall \\
    \midrule
    %SVM  & 0.816 & 0.812 & 0.809 & 0.822& 0.964 & 0.871 & 0.855 & 0.894 \\
    %LR  & 0.822 & 0.815 & 0.812 & 0.819& 0.963 & 0.874 & 0.862 & 0.891 \\
    %DT  & 0.871 & 0.868 & 0.864 & 0.876&0.950 & 0.849 & 0.842 & 0.858 \\
    %RF  & 0.807 & 0.774 & 0.852 & 0.760& 0.957 & 0.891 & 0.909 & 0.877 \\
    %BiGRU-100d (N+C+E) (7M) & 0.580 & 0.367 & 0.290 & 0.500& 0.636 & 0.421 & 0.363 & 0.500 \\
    BiGRU (N+C+E) (28M) & 0.580 & 0.367 & 0.290 & 0.500& 0.876 & 0.782 & 0.769 & 0.804 \\
    %BERT (N) (110M) & 0.703 & 0.659 & 0.729 & 0.662 & 0.909& 0.799& 0.791 &0.810 \\
    %BERT (C) (110M) &0.777&0.770&0.772&0.768&0.895&0.813&0.798&0.846\\
    %BERT (news+desc)& 0.670 & 0.628 & 0.674 & 0.632 & 0.915& 0.803& 0.794&0.814\\
    %BERT (N+C)&0.722&0.693&0.736&0.690&0.911&0.872&0.890&0.859\\
    BERT (N+C+E) (339M / 110M) &0.787&0.776&0.787&0.771 &0.940&0.877&0.901&0.859\\
    %RoBERTa (N) (125M) &0.715&0.670&0.754&0.673&0.923&0.799&0.791&0.810\\
    %RoBERTa (C) (125M) &0.846&0.845&0.843&0.851&0.838&0.813&0.798&0.846\\
    %RoBERTa (news+desc)&0.700&0.676&0.699&0.673&0.912&0.799&0.791&0.810\\
    %RoBERTa (N+C)&0.735&0.703&0.760&0.701&\textbf{0.962}&0.877&0.901&0.859\\
    RoBERTa (N+C+E) (384M / 125M) &0.894&0.890&0.896&0.887&0.918&0.877&0.901&0.859\\
    %LinkBERT (N) (110M) &0.707&0.669&0.726&0.670&0.928&0.799&0.791&0.810\\
    %LinkBERT (C) (110M) &0.809&0.796&0.819&0.789&0.918&0.813&0.798&0.846\\
    %LinkBERT (news+desc)&0.719&0.695&0.722&0.692&0.895&0.800&0.791&0.813\\
    %LinkBERT(N+C)&0.749&0.735&0.748&0.730&0.923&0.878&0.898&0.862\\
    LinkBERT (N+C+E) (330M / 110M) & 0.824 &0.811&0.841&0.802&0.927&0.880&0.903&0.863\\
    dEFEND (N+C) (5M) & 0.771 & 0.758 & 0.771 & 0.754& 0.749 & 0.799 & 0.792 & 0.808 \\
    % dEFEND-300d (N+C) (16M) &0.634&0.535&0.673&0.574&0.766&0.804&0.789&0.840\\
    \midrule
    %Dual-CAN-100d (N+C+E) (8M) &0.927 & 0.926 &0.924 &0.927 &0.937 & 0.807 & 0.797 &0.819\\
    Dual-CAN (N+E) (33M) &0.895&0.891&0.901&0.885&0.853&0.884&0.905&0.868\\
    Dual-CAN (N+C) (33M) &0.914&0.912&0.913&0.911&0.937&\textbf{0.887}&\textbf{0.907}&\textbf{0.872}\\
    Dual-CAN (N+C+E) (33M) &\textbf{0.949} & \textbf{0.947} &\textbf{0.946} &\textbf{0.949} & \textbf{0.954} & 0.884 & 0.905 & 0.868\\
    \bottomrule
    \end{tabular}
    \caption{Experimental results. N, C, and E denote news content, user comments, and entity description, respectively. BERT-based models are implemented in two methods (details in Appendix \ref{sec:Implement Detail}) with different number of parameters. }
    \label{Table result}
\end{table*}

\subsection{Prediction Component}
\label{sec:4.5}
Our task is a binary classification task with real/fake labels.
First, we concatenate all feature vectors $\mathbf{f}=[\hat{\mathbf{s_1}}, \hat{\mathbf{d}}, \hat{\mathbf{s_2}}, \hat{\mathbf{c}}]$, and feed the result into a 2-layer linear neural network. 
It is calculated by:
\begin{equation}
    \mathbf{\hat{y}}=\mathbf{W_2}(\mathbf{W_1f}+\mathbf{b_1})+\mathbf{b_2}
\end{equation}
where $\mathbf{W_1}$ and $\mathbf{W_2}$ are learnable parameters and $\mathbf{b_1}, \mathbf{b_2}$ are bias terms. The prediction result $\mathbf{\hat{y}}=[y_0,y_1]$ indicates the probabilities of label 0 is $y_0$, and label 1 is $y_1$. 
We choose cross entropy as our loss function:
\begin{equation}
    \mathcal{L}(\theta)=-ylog(\hat{y_1})-(1-y)log(1-\hat{y_0})
\end{equation}
where $\theta$ is all parameters in our model. We choose Adam optimizer \cite{https://doi.org/10.48550/arxiv.1412.6980} to optimize all parameters $\theta$.

\section{Experiments}
\subsection{Datasets}
We adopt two datasets in our experiment. %\footnote{Please download the dataset via the link in Appendix \ref{sec:dataset download}.}
The first dataset is GossipCop \cite{https://doi.org/10.48550/arxiv.1809.01286}, which collects both news content and social context from fact-checking website.
The second dataset is CoAID \cite{https://doi.org/10.48550/arxiv.2006.00885}, which is a benchmark dataset for COVID-19 misinformation. 
Please refer to Appendix \ref{sec:dataset download} for the statistics of the datasets.
We follow the evaluation settings as previous studies~\cite{https://doi.org/10.48550/arxiv.1809.01286,https://doi.org/10.48550/arxiv.2006.00885} to use (Accuracy, F1, Precision, Recall) for \emph{GossipCop} and use (PR-AUC, F1, Precision, Recall) for \emph{CoAID}.

\subsection{Results}
\label{sec: results}
We compare the results with the following representative models: 
BiGRU~\cite{https://doi.org/10.48550/arxiv.1412.3555}, 
BERT~\cite{devlin-etal-2019-bert}, 
RoBERTa~\cite{liu2019roberta}, 
LinkBERT~\cite{yasunaga-etal-2022-linkbert}), and 
dEFEND~\cite{shu2019defend}.\footnote{Because \citet{shu2019defend} did not release the information for dataset separation, we use the same hyperparameter reported in their work to reproduce the results. All implemental details are provide in Appendix \ref{sec:Implement Detail}}
Table~\ref{Table result} shows our experimental results. 
Our Dual-CAN outperforms all baselines in both datasets. 
In addition, our Dual-CAN uses fewer parameters than BERT-based models.
Our approach also performs better than dEFEND \cite{shu2019defend} when no entity descriptions are provided. This is because we use different preprocessing methods, and the differences between two model architectures. 
The bottom half of Table~\ref{Table result} shows ablation analysis of the proposed model. 
The results indicate the importance of adding entity information to the proposed model, especially in \textit{GossipCop}. 
However, only a few improvements in PR-AUC when experimenting with \textit{CoAID}.  \textit{CoAID} usually are short posts that contain few entities, which results in the limitation of the proposed entity-aware concept. The main source to predict whether a piece of news is fake is the news content itself. Therefore, N+E, N+C, and N+C+E results only have small differences because they both contain N. The roles of C and E are to improve the predictions.

\begin{figure}
\centering
\resizebox{4.5cm}{!}{%
    \begin{subfigure}[b]{.24\linewidth}
    \centering
    \includegraphics[width=\linewidth]{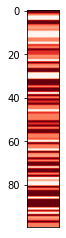}
    \caption{}\label{fig1a}
    \end{subfigure}\hfill
    \begin{subfigure}[b]{.24\linewidth}
    \centering
    \includegraphics[width=\linewidth]{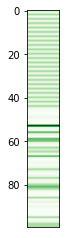}
    \caption{}\label{fig1b}
    \end{subfigure}\hfill
    \begin{subfigure}[b]{.24\linewidth}
    \centering
    \includegraphics[width=\linewidth]{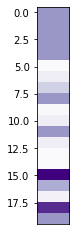}
    \caption{}\label{fig1c}
    \end{subfigure}\hfill
    \begin{subfigure}[b]{.24\linewidth}
    \centering
    \includegraphics[width=\linewidth]{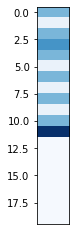}
    \caption{}\label{fig1d}
    \end{subfigure}%
}
    \caption{Attention weights of: (a) \emph{GossipCop} entity description, (b) \emph{GossipCop} user comments, (c) \emph{CoAID} entity description, and (d) \emph{CoAID} user comments. Dark colors means higher attention weights. The vertical axis means the index of the sentence.}
    \label{fig:attention weight}
\end{figure}

\section{Interpretability}
\label{sec:Model Interpretability Evaluation}
We examine attention weights $[\mathbf{a_{s_1}},\mathbf{a_{d}},\mathbf{a_{s_2}},\mathbf{a_{s_c}}]$ to find those sentences that the proposed model is focusing on when making predictions. 
Figure~\ref{fig:attention weight} illustrates the results.
We find that our model pays a certain degree of attention to the first sentence in the entity descriptions of both datasets (Figure~\ref{fig1a},\ref{fig1c}). 
Our intuition about this phenomenon is that the first sentence always provides a brief definition of the entity, and it would be helpful for models to understand the given entity. 
%In contrast, for \textit{CoAID} instances (Figure~\ref{fig1c}), model pays attention to the later sentences.%
%Since CoAID's news content is shorter, entities that extract first may be more general entities. Therefore, the later entities descriptions are more important for the models. 
On the other hand, model's attention weights on user comments of both datasets are in the middle replies, as shown in Figure~\ref{fig1b} and Figure~\ref{fig1d}. 
It follows our intuition because the sentences like ``FYI. It's a fake news.'' for clarifying the given news/post is fake news always appears later than some discussions. 
Based on Figure~\ref{fig1d}, we also find that models give little attention weight to the twelfth or later sentences.
Besides weight distributions studies, we also did some case studies in Appendix \ref{sec:case study}. The results show that attention weights do reflect the important parts of the input, which help us interpret the model better. For example, we understood the importance and usage of entity descriptions from attention weights.  

\section{Conclusion}
We propose a dual co-attention network for fake news detection, which improves the previous representative model, dEFEND, by (1) adding entity description as external knowledge and (2) redesigning co-attention architecture for using all input information. 
Our results support the usefulness of the proposed Dual-CAN model. 
The interpretability based on the attention weight is also discussed.

\appendix
\section*{Limitations}
The major limitation of the proposed model is that when the given text (news article or social media post) is short, and the performance of adding entity description may not be significantly improved. 
It is because such text provides few entities in the narrative, and it will limit the proposed entity-aware concept.

\section*{Ethical Statement}
We will follow the licenses of GossipCop \cite{https://doi.org/10.48550/arxiv.1809.01286} and CoAID \cite{https://doi.org/10.48550/arxiv.2006.00885} to share the training, development, and test datsets in our experiments.

\section*{Acknowledgments}
This research is supported by National Science and Technology Council, Taiwan, under grants 110-2221-E-002-128-MY3, 110-2634-F-002-050-, and 111-2634-F-002-023-.

% Entries for the entire Anthology, followed by custom entries
\bibliographystyle{acl_natbib}
\bibliography{custom}

\appendix

\begin{table}
    \centering
    \resizebox{\columnwidth}{!}{%
        \begin{tabular}{l|c|c}
        \hline
        \textbf{Datasets} & \textbf{GossipCop} &\textbf{CoAID}\\
        \hline
        Word Embedding & \begin{tabular}{c|c}\makecell{Glove \\100d}  &  \makecell{Glove \\300d}\\\end{tabular} &  \begin{tabular}{c|c}\makecell{Glove \\100d}  &  \makecell{Glove \\300d}\\\end{tabular} \\
        \hline
        Max sentence length & 120 &120 \\
        \hline
        \makecell[l]{Max sentence number \\ per news $N$} & 40 & 4 \\
        \hline
        \makecell[l]{Max sentence number \\ per entity description} & 4 & 4 \\
        \hline
        \makecell[l]{Max sentence number \\ per user comment} & 2 & 2 \\
        \hline
        \makecell[l]{Max sentence number \\ of total entity description $E$} & 100 & 20 \\
        \hline
        \makecell[l]{Max sentence number \\ of total user comment $U$} & 100 & 20 \\
        \hline
        Embedding dimension & \begin{tabular}{c|c}100  &  300\\\end{tabular} &  \begin{tabular}{c|c}100  &  300\\\end{tabular} \\
        \hline
        h & \begin{tabular}{c|c}100  &  300\\\end{tabular} & \begin{tabular}{c|c}100  &  300\\\end{tabular} \\
        \hline
        Batch size & 16 & 32 \\
        \hline
        learning rate & 0.001 & 0.001 \\
        \hline
        \end{tabular}
    }
    \caption{Model parameters.}
    \label{Table param}
\end{table}

\section{Dataset}
\label{sec:dataset download}
Table~\ref{Table stat} reports the statistics of the datasets.
We will release the datasets for reproduction, and follow the same license of GossipCop and CoAID.
Due to the size limits, we cannot upload the dataset via the submission system. Please download it via the following anonymous link: \url{https://drive.google.com/file/d/1QuZeINFHqy8OF1Av5627zTyyVVg7g2HD/view?usp=sharing}.

\begin{table}
   \small
    \centering
    \begin{tabular}{lrr}
    \toprule
    \textbf{Datasets} & \textbf{GossipCop} &\textbf{CoAID}\\
    \midrule
    Total news  & 4,273 &  2,162 \\    
    True news  & 2,562 & 1,590  \\
    Fake news  & 1,711 & 572  \\
    User Comments  & 309,059 & 37,187  \\
    Entity Descriptions  & 95,150 & 5,666  \\
    \bottomrule
    \end{tabular}
    \caption{Dataset statistics.}
    \label{Table stat}
\end{table}

\section{Implementation Detail}
\label{sec:Implement Detail}
Below are the implementation details of the baseline models:
\begin{itemize}
    %\item \textbf{SVM}: We use bag of words for preprocessing, and feed into linear SVM.
    %\item \textbf{Logistic Regression}: We use bag of words for preprocessing, and feed into the model.
    %\item \textbf{Decision Tree}: We use bag of words for preprocessing, and feed into the model.
    %\item \textbf{Random Forests}: We use bag of words for preprocessing, and feed into the model.
    \item \textbf{BiGRU} \cite{https://doi.org/10.48550/arxiv.1412.3555}: We use Glove 300d for word embedding of news content, entity descriptions and user comments. The word embedding of three resources are feed into BiGRU and concatenate their results $\mathbf{T}=[v_n,v_d,v_c]$. 
    Second, we feed $\mathbf{T}$ into linear neural network described in Section \ref{sec:4.5} to get final result.
    
    \item Pretrained language models (\textbf{BERT} \cite{devlin-etal-2019-bert}, \textbf{RoBERTa} \cite{liu2019roberta}, \textbf{LinkBERT} \cite{yasunaga-etal-2022-linkbert}): We adopt three representative pretrained language models for comparison, and implemented in two different ways.\\
    \begin{itemize}
        \item \textbf{Method 1}\\
        For each experiment, we feed news content, entity descriptions, and user comments into three tokenizers respectively. Afterwards, we feed input id and attention masks of each resource into three pretrained language models respectively. Each pretrained model handles one resource. Finally, we concatenate the outputs of three pretrained models, and pass a linear layer to output probability of two labels $\hat{y}$. This method is used for \emph{GossipCop} dataset in Table \ref{Table result}.
        \item \textbf{Method 2}\\
        The number of parameters in \textbf{Method 1} is huge, but it's necessary for \emph{GossipCop} dataset. We tried another method to reduce the number of parameters. We concatenate all three resources and feed into one tokenizer. Second, we feed input id and attention masks into one pretrained models. The final procedure is same as the previous method. The experiment results for \emph{CoAID} dataset are in Table \ref{Table result}, and the results are better than \textbf{Method 1}'s. The experiment results for \emph{GossipCop} dataset is in Table \ref{Table result for BERT}. The performances are worse than \textbf{Method 1}'s. We believe it's because \emph{GossipCop} dataset's data are too long for a single pretrained model. Therefore, we tried \textbf{Longformer} \cite{beltagy2020longformer} which accept longer input. The performance becomes better, but this methods uses more parameters.
    \end{itemize}
    
    \item \textbf{dEFEND} \cite{shu2019defend}: dEFEND is one of the representative fake news detection methods. It is based on co-attention model to increase explainability.\footnote{Because \citet{shu2019defend} did not release the information for dataset separation, we use the same hyperparameter reported in their work to reproduce the results. We will release the datasets for reproduction.} 

\end{itemize}

\begin{table*}[t]
    \centering
    \small
    \begin{tabular}{lcccc}
    \toprule
    \multirow{2}[2]{*}{Model (Input) (\# of Parameters)} & \multicolumn{4}{c}{GossipCop}\\
     & Accuracy & F1 & Precision & Recall \\
    \midrule
    BERT (N+C+E) (110M) &0.643&0.587&0.645&0.599 \\
    RoBERTa (N+C+E) (125M) &0.698&0.631&0.774&0.646\\
    LinkBERT (N+C+E) (110M) & 0.702&0.694&0.694&0.693\\
    Longformer (N+C+E) (148M) & 0.752 &0.742&0.758 & 0.723\\
    \bottomrule
    \end{tabular}
    \caption{\textbf{Method 2} experiment results of GossipCop dataset. N, C, and E denote news content, user comments, and entity description, respectively.}
    \label{Table result for BERT}
\end{table*}

Table~\ref{Table param} reports the hyperparameters used in the proposed Dual-CAN. In the ablation study, we remove the original data of entity description E or user comments C, and replace them with padding token <PAD>. Therefore, the model architecture remains the same as Section \ref{sec:method} stated. We have submitted the code for review, and it will be released on GitHub.

\section{Case Study of Interpretability}
\label{sec:case study}
We analyzed individual sentences and words which have higher attention weight, in order to figure out the explanability of the attention weight. 

For sentence-level analysis, entity descriptions that define an entity would have higher attention weights. Here are two example entity descriptions that have higher attention weights:
\begin{enumerate}
    \item \textbf{\{Dataset: GossipCop, id: 587, attention weight: 0.036 >average 0.01\}}:
\emph{“IMDb (an abbreviation of Internet Movie Database) is an online database of information related to films, television series, home videos,…”}
    \item \label{case study 2} \textbf{\{Dataset: CoAID, id: 48, attention weight: 0.182 >average 0.05\}}:
\emph{“The Centers for Disease Control and Prevention (\textbf{CDC}) is the national public health agency of the United States.”}
    \item \label{case study 3} \textbf{\{Dataset: CoAID, id: 1304, attention weight: 0.119 > average 0.05\}}: \emph{“\textbf{Getty Images}, Inc. is a British-American visual media company and is a supplier of stock images, editorial photography, video and music for business and consumers, with a library of over 477 million assets.”}
\end{enumerate}

Moreover, we can see some correlation between highlighted entity descriptions and news content that contain them. For example, the news sentence which contains entity (\ref{case study 2},\ref{case study 3}), both have higher attention weight than average.

\begin{enumerate}
    \item \textbf{\{Dataset: CoAID, id: 48, attention weight: 0.33 >average 0.25\}}:
\emph{“enters for disease control and prevention, \textbf{cdc} twenty four seven, saving lives protecting people centers for disease control and prevention”}
    \item \textbf{\{Dataset: CoAID, id: 1304, attention weight: 0.33 > average 0.25\}}: \emph{“\textbf{getty images} the antimalarial drug hydroxychloroquine is being widely promoted as a cure for covid-19 but we still lack good data on its true benefits.”}
\end{enumerate}

Case studies indicate that our model performs like it is doing “fact-checking”, which is an useful and important strategy for fake news detection. Meanwhile, entity descriptions are essential for fact-checking. Therefore, with the good usage of entity descriptions, fake news detection can achieve better performance, same as the ablation studies in  Section \ref{sec: results} shown.

For word-level analysis, we discovered similar results as  \cite{lu-li-2020-gcan} did. Some fake news contains emotional words or words that catch people’s attention like “Breaking”or “warn”.

\end{document}